\pdfoutput=1

\documentclass[11pt]{article}

\usepackage{acl}

\usepackage{times}
\usepackage{latexsym}

\usepackage[T1]{fontenc}

\usepackage[utf8]{inputenc}

\usepackage{microtype}

\usepackage{CJKutf8}
\usepackage{xcolor}
\usepackage{hyperref}
\usepackage{graphicx}
\usepackage{paralist}
\usepackage{eucal}
\usepackage{amsmath}
\usepackage{amssymb}
\usepackage{multirow}
\usepackage{xspace}
\usepackage[switch]{lineno}  

\title{Rethinking Document-level Neural Machine Translation}

\author{
Zewei Sun\textsuperscript{\rm 1,2,\thanks{* Work was done while at ByteDance}}, 
Mingxuan Wang\textsuperscript{\rm 2},
Hao Zhou\textsuperscript{\rm 2},
Chengqi Zhao\textsuperscript{\rm 2} \\
\textbf{Shujian Huang}\textsuperscript{\rm 1,3},
\textbf{Jiajun Chen}\textsuperscript{\rm 1},
\textbf{Lei Li}\textsuperscript{\rm 4,*}\\
\textsuperscript{\rm 1} State Key Laboratory for Novel Software Technology, Nanjing University \\ 
\textsuperscript{\rm 2} ByteDance AI Lab, \textsuperscript{\rm 3} Peng Cheng Laboratory, Shenzhen \\
\textsuperscript{\rm 4} University of California, Santa Barbara\\
\texttt{\{sunzewei.v,wangmingxuan.89\}@bytedance.com} \\ 
\texttt{\{zhouhao.nlp,zhaochengqi.d\}@bytedance.com} \\ 
\texttt{\{huangsj,chenjj\}@nju.edu.cn, lilei@cs.ucsb.edu}
}

\begin{document}
\maketitle
\begin{abstract}

This paper does not aim at introducing a novel model for document-level neural machine translation. Instead, we head back to the original Transformer model and hope to answer the following question: Is the capacity of current models strong enough for document-level translation? Interestingly, we observe that the original Transformer with appropriate training techniques can achieve strong results for document translation, even with a length of 2000 words. We evaluate this model and several recent approaches on nine document-level datasets and two sentence-level datasets across six languages. Experiments show that document-level Transformer models outperforms sentence-level ones and many previous methods in a comprehensive set of metrics, including BLEU, four lexical indices, three newly proposed assistant linguistic indicators, and human evaluation.

\end{abstract}

\section{Introduction}
\label{sec:intro}


Neural machine translation~\cite{Bahdanau2015NeuralMT,Wu2016GooglesNM,Vaswani2017AttentionIA} has achieved great progress and reached near human-level performance. However, most current sequence-to-sequence NMT models translate sentences individually.
In such cases, discourse phenomena, such as pronominal anaphora, lexical consistency, and document coherence that depend on long-range context going further than a few previous sentences, are neglected~\cite{Bawden2017EvaluatingDP}. As a result, \citet{laubli2018has} find human raters still show a markedly stronger preference for human translations when evaluating at the level of documents. 

Many methods have been proposed to improve document-level neural machine translation (DNMT). Among them, the mainstream studies focus on the model architecture modification, including hierarchical attention~\cite{Wang2017ExploitingCC,Werlen2018DocumentLevelNM,Tan2019HierarchicalMO}, additional context extraction encoders or query layers~\cite{Jean2017NeuralMT,Bawden2017EvaluatingDP,Zhang2018ImprovingTT,Voita2018ContextAwareNM,Kuang2018FusingRI,Maruf2019SelectiveAF,Yang2019EnhancingCM,Jiang2019DocumentlevelNM,zheng2020toward,yun2020improving,xuefficient}, and cache-like memory network~\cite{Maruf2018DocumentCN,Kuang2018ModelingCF,Tu2018LearningTR}.

These studies come up with different structures in order to include discourse information, namely, introducing adjacent sentences into the encoder or decoder as document contexts. Experimental results show effective improvements on universal translation metrics like BLEU~\cite{Papineni2001BleuAM} and document-level linguistic indices~\cite{Tiedemann2017NeuralMT,Bawden2017EvaluatingDP,Werlen2017ValidationOA,Mller2018ALT,Voita2018ContextAwareNM,Voita2019WhenAG}.

Unlike previous work, this paper does not aim at introducing a novel model. Instead, we hope to answer the following question: Is the basic sequence-to-sequence model strong enough to directly handle document-level translation? To this end, we head back to the original Transformer~\cite{Vaswani2017AttentionIA} and conduct literal document-to-document (Doc2Doc) training. 

Though many studies report negative results of naive Doc2Doc translation~\cite{Zhang2018ImprovingTT,Liu2020MultilingualDP}, we successfully activate it with \textit{Multi-resolutional Training}, which involves multiple levels of sequences.
It turns out that end-to-end document translation is not only feasible but also stronger than sentence-level models and previous studies. Furthermore, if assisted by extra sentence-level corpus, which can be much more easily obtained, the model can significantly improve the performance and achieve state-of-the-art results. It is worth noting that our method does not change the model architecture and need no extra parameters.

Our experiments are conducted on nine document-level datasets, including TED (ZH-EN, EN-DE), News (EN-DE, ES-EN, FR-EN, RU-EN), Europarl (EN-DE), Subtitles (EN-RU), and a newly constructed News dataset (ZH-EN). Additionally, two sentence-level datasets are adopted in further experiments, including Wikipedia (EN-DE) and WMT (ZH-EN).
Experiment results show that our strategy outperforms previous methods in a comprehensive set of metrics, including BLEU, four lexical indices, three newly proposed assistant linguistic indicators, and human evaluation. In addition to serving as improvement evidence, our newly proposed document-level datasets and metrics can also be a boosting contribution to the community.

\section{Re-examining Recent DNMT Studies}
\label{sec:reg}

For DNMT, though many improvements have been reported, a couple of studies have proposed challenges against these results~\cite{Kim2019WhenAW,Jwalapuram2020CanYC,li2020does}. And we also find some of previous gains should be attributed to overfitting to some extent.

The most used datasets of previous work are \textit{News Commentary} and \textit{TED Talks}, which contain only 200 thousand sentences. 
The small scale of the datasets gives rise to the frequent occurrence of overfitting, let alone that the distribution of the test set is highly similar to the training set.
And some work even conduct an unfair comparison with dropout=0.1 for sentence-level models and dropout=0.2 for document-level models ~\cite{Maruf2019SelectiveAF,Yang2019EnhancingCM,zheng2020toward}. 
As a result, the parameter regularization and overfitting on small datasets make the improvements not solid enough.



To verify our assumption,
we perform different training by switching hyperparameters on sentence-level experiments. We follow the datasets provided by \citet{Maruf2019SelectiveAF} and \citet{zheng2020toward}, including \textit{TED} (ZH-EN/EN-DE), \textit{News} (EN-DE), and \textit{Europarl} (EN-DE), as well as all the model architecture settings they adopt, including a four-layer Transformer base version.

\begin{table}[ht]\scriptsize
    \centering
    \begin{tabular}{l|c|ccc}
        \hline
        \multirow{2}{*}{Models} & ZH-EN & \multicolumn{3}{c}{EN-DE} \\
         & TED & TED & News & Europarl \\ \hline
        Transformer-base (dropout=0.1) & 17.32 & 23.58 & 22.10 & \textbf{31.70} \\
        Transformer-base (dropout=0.2) & 18.87 & 24.70 & 24.36 & 31.44 \\
        Transformer-base (dropout=0.3) & \textbf{19.21} & \textbf{25.19} & 24.98 & 30.56 \\
        \hline
        DocT~\cite{Zhang2018ImprovingTT} & - & 24.00 & 23.08 & 29.32 \\
        HAN~\cite{Werlen2018DocumentLevelNM} & 17.90 & 24.58 & \textbf{25.03} & 28.60 \\
        SAN~\cite{Maruf2019SelectiveAF} & - & 24.42 & 24.84 & 29.75 \\
        QCN~\cite{Yang2019EnhancingCM} & - & 25.19 & 22.37 & 29.82 \\
        MCN~\cite{zheng2020toward} & 19.10 & 25.10 & 24.91 & 30.40 \\ \hline
    \end{tabular}
    \captionsetup{font={footnotesize}}
    \caption{Document translation experiments on ZH-EN and EN-DE. ``-'' means not provided. Only the results of TED \& News with dropout=0.1 and a much lower score of Europarl are reported in previous work. However, Transformer-base with dropout=0.3 for TED \& News and a strong baseline of Europarl outperform almost all other methods.}
    \label{tab:reg}
\end{table}

As is shown in Table~\ref{tab:reg}, we surprisingly find that simply employing larger dropout can eliminate all the improvements gained by previous work. 
For \textit{TED}, the setting of dropout=0.2 can boost baseline for more than 1.0 BLEU, which immediately marginalizes the previous advance, while the setting of dropout=0.3 can outperform all the previous studies. 
When it comes to \textit{News}, though the state-of-the-art results are yet to be obtained, the gap between sentence and document models has been largely narrowed up.
As for \textit{Europarl}, a much higher baseline has been easily achieved, which also makes other improvements not solid enough.

Our results show that preceding experiments lack the comparison with a strong baseline. An important proportion of the improvements may come from the regularization of the models since they bring in extra parameters for context encoders or hierarchical attention weights. However, the regularization can be also achieved in sentence-level models and is not targeted at improving document coherence. Essentially, the small scale of related datasets and identically distributed test sets make the improvements questionable.

\citet{Kim2019WhenAW} draw the same conclusion that well-regularized or pre-trained sentence-level models can beat document-level models in the same settings. They check the translation and find that most improvements are not from coreference or lexical choice but ``not interpretable". Similarly, \citet{Jwalapuram2020CanYC} adopt a wide evaluation and find that the existing context-aware models do not improve discourse-related translations consistently across languages and phenomena. Also, \citet{li2020does} find that the extra context encoders act more like a noise generator and the
BLEU improvements mainly come from the robust training instead of the leverage of contextual information. All these three studies appeal for stronger baselines for a fair comparison.

We suggest that the current research tendency in DNMT should be reviewed since it is hard to tell whether the improvements are targeted at document coherence or just normal regularization, let alone complicated modules are introduced. Therefore, as a simpler alternative, we head back to the original but concise style, using end-to-end training framework to cope with document translation.

\section{Doc2Doc: End-to-End DNMT}
\label{sec:context}

In this section, we attempt to analyze the different training patterns for DNMT. Firstly, let us formulate the problem. 
Let $D_x=\{x^{(1)}, x^{(2)},\cdots, x^{(M)}\}$ be a source-language document containing $M$ source sentences. 
The goal of the document-level NMT is to translate the document $D_x$ in language $x$ to a document $D_y$ in language $y$. $D_y=\{y^{(1)}, y^{(2)},\cdots, y^{(N)}\}$.
We use $L_y^{(i)}$ to denote the sentence length of $y^{(i)}$. 

Previous work translate a document sentence-by-sentence, regarding DNMT as a step-by-step sentence generating problem (Doc2Sent) as:

\begin{equation}\small
      \mathcal{L}_\text{Doc2Sent}= -\sum_{i=1}^{N}\sum_{j=1}^{L_y^{(i)}} \log p_\theta(y_j^{(i)}|y_{(<j)}^{(i)},x^{(i)},S^{(i)},T^{(i)}),
      \label{eq:dsneglike}
\end{equation}

$S^{(i)}$ is the context in the source side, depending on the model architecture and is comprised of only two or three sentences in many work. Most current work focus on $S^{(i)}$, by utilizing hierarchical attention or extra encoders. And $T^{(i)}$ is the context in the target side, which is involved by only a couple of work. They usually make use of a topic model or word cache to form $T^{(i)}$.

Different from Doc2Sent, we propose to resolve document translation with the end-to-end, namely document-to-document (Doc2Doc) pattern as:

\begin{equation}\small
      \mathcal{L}_\text{Doc2Doc}= -\sum_{i=1}^{\sum_{L_y}} \log p_\theta(y_i|y_{<i},D_x),
      \label{eq:dneglike}
\end{equation}

where $D_x$ is the complete context in the source side, and $y_{<i}$ is the complete historical context in the target side.

\subsection{Why We Dive into Doc2Doc?}
\label{subsec:why}

\paragraph{Full Source Context:} 
First, many Doc2sent studies show that more sentences beyond can harm the results~\cite{Werlen2018DocumentLevelNM,Zhang2018ImprovingTT,Tu2018LearningTR}. Therefore, many Doc2Sent work are more of ``a couple of sentences to sentence'' since they only involve two or three preceding sentences as context.
However, broader contexts provide more information, which shall theoretically lead to more improvements.
Thus, We attempt to re-visit involving the full context and choose Doc2Doc, as it is required to take account of all the source-side context.

\paragraph{Full Target Context:} 
Second, many Doc2sent work abandon the target-side historical context, and some even claim that it is harmful to translation quality~\cite{Wang2017ExploitingCC,Zhang2018ImprovingTT,Tu2018LearningTR}.
However, once the cross-sentence language model is discarded, some problems, such as tense mismatch (especially when the source language is tenseless like Chinese), may occur. 
Therefore, we attempt to re-visit involving the full context and choose Doc2Doc, as it treats the whole document as a sequence and can naturally take advantage of all the target-side historical context.

\paragraph{Relaxed Training:} Third, Doc2Sent restricts the training scene. The previous work focus on adjusting the model structure to feed preceding source sentences, so the training data has to be in the form of consecutive sentences so as to meet the model entrance. As a result, it is hard to use large numbers of piecemeal parallel sentences.
Such a rigid form of training data also greatly limits the model potential because the scale of parallel sentences can be tens of times of parallel documents. On the contrary, Doc2Doc can naturally absorb all kinds of sequences, including sentences and documents.

\paragraph{Simplicity:} Last, Doc2Sent inevitably introduces extra model modules with extra parameters in order to capture contextual information. It complicates the model architecture, making it hard to renovate or generalize. On the contrary, Doc2Doc does not change the model structure and brings in no additional parameters.

\begin{table*}[htb]\scriptsize
    \centering
    \begin{tabular}{cccccccc}
        \hline
        Group & Datasets & Source & Language & N\_Sent & N\_Doc & Development Sets & Test Sets \\ \hline
        \multirow{4}{*}{Main Experiments} & \href{https://wit3.fbk.eu/mt.php?release=2015-01}{TED} & IWSLT 2015 & ZH-EN & 205K & 1.7K & dev2010 & tst2010-2013 \\
         & \href{https://github.com/sameenmaruf/selective-attn}{TED} & IWSLT 2017 & EN-DE & 206K & 1.7K & dev2010+tst201[0-5] & tst2016-2017 \\
         & \href{https://github.com/sameenmaruf/selective-attn}{News} & News Commentary v11 & EN-DE & 236K & 6.1K & newstest2015 & newstest2016 \\
         & \href{https://github.com/sameenmaruf/selective-attn}{Europarl} & Europarl v7 & EN-DE  & 1.67M & 118K & \multicolumn{2}{c}{\cite{Maruf2019SelectiveAF}} \\ \hline
        \multirow{3}{*}{Other Languages} & \href{http://data.statmt.org/news-commentary/v14/}{News} & News Commentary v14 & ES-EN & 355K & 9.2K & newstest2012 & newstest2013 \\
         & \href{http://data.statmt.org/news-commentary/v14/}{News} & News Commentary v14 & FR-EN & 303K & 7.8K & newstest2013 & newstest2014 \\
         & \href{http://data.statmt.org/news-commentary/v14/}{News} & News Commentary v14 & RU-EN & 226K & 6.0K & newstest2018 & newstest2019 \\ \hline
        \multirow{2}{*}{Sentence-level Corpus} & \href{http://opus.nlpl.eu/Wikipedia.php}{Wiki} & Wikipedia & EN-DE & 2.40M & - & - & - \\
         & \href{http://www.statmt.org/wmt19/translation-task.html}{WMT} & WMT 2019 & ZH-EN & 21M & - & - & - \\ \hline
        Contrastive Experiments & \href{https://github.com/lena-voita/good-translation-wrong-in-context#training-data}{Subtitles} & OpenSubtitles & EN-RU & 6M & 1.5M & \multicolumn{2}{c}{\cite{Voita2019WhenAG}} \\ \hline 
        \multirow{1}{*}{Our New Datasets} & \href{https://github.com/sunzewei2715/Doc2Doc_NMT}{PDC} & FT/NYT & ZH-EN & 1.39M & 59K & newstest2019 & PDC \\
        \hline
    \end{tabular}
    \caption{The detailed information of the used datasets in this paper with downloading links on their names.}
    \label{tab:datasets}
\end{table*}

\subsection{Multi-resolutional Doc2Doc NMT}

Although Doc2Doc seems more concise and promising in multiple terms, it is not widely recognized. \citet{Zhang2018ImprovingTT,Liu2020MultilingualDP} conduct experiments by directly feeding the whole documents into the model. We refer to it as \textit{Single-resolutional Training} (denoted as SR Doc2Doc). Their experiments report extremely negative results unless pre-trained in advance. The model either has a large drop in performance or does not work at all. As pointed out by ~\citet{Koehn2017SixCF}, one of the six challenges in neural machine translation is the dramatic drop of quality as the length of the sentences increases.
 
However, we find that Doc2Doc can be activated on any datasets and obtain better results than Doc2Sent models as long as we employ \textit{Multi-resolutional Training}, mixing documents with shorter segments like sentences or paragraphs (denoted as MR Doc2Doc). 

Specifically, we split each document averagely into $k$ parts for multiple times and collect all the sequences together, $k \in \{1,2,4,8,...\}$. For example, a document containing eight sentences will be split into two four-sentences segments, four two-sentences segments, and eight single-sentence segments. Finally, fifteen sequences are all gathered and fed into sequence-to-sequence training $(15=1+2+4+8)$. 

In this way, the model can acquire the ability to translate long documents since it is assisted by easier and shorter segments. As a result, multi-resolutional Doc2Doc is able to translate all forms of sequences, including extremely long ones such as a document with more than 2000 tokens, as well as shorter ones like sentences. In the following sections, we conduct the same experiments as the aforementioned studies by translating the whole document directly and atomically.

\begin{table*}[ht]\footnotesize
    \centering
    \resizebox{\textwidth}{32mm}{
    \begin{tabular}{l|cc|cccccc}
        \hline
        \multirow{3}{*}{Models} & \multicolumn{2}{|c|}{ZH-EN} & \multicolumn{6}{c}{EN-DE} \\
         & \multicolumn{2}{|c|}{TED} & \multicolumn{2}{c}{TED} & \multicolumn{2}{c}{News} & \multicolumn{2}{c}{Europarl} \\
         & s-BLEU & d-BLEU & s-BLEU & d-BLEU & s-BLEU & d-BLEU & s-BLEU & d-BLEU \\ \hline
        Sent2Sent~\cite{zheng2020toward} & 17.0 & - & 23.10 & - & 22.40 & - & 29.40 & - \\
        Sent2Sent (Our strong baseline) & 19.2 & 25.8 & 25.19 & 29.16 & 24.98 & 27.03 & 31.70 & 33.83 \\ \hline
        DocT~\cite{Zhang2018ImprovingTT} & - & - & 24.00 & - & 23.08 & - & 29.32 & - \\
        HAN~\cite{Werlen2018DocumentLevelNM} & 17.9 & - & 24.58 & - & 25.03 & - & 28.60 & -\\
        SAN~\cite{Maruf2019SelectiveAF} & - & - & 24.42 & - & 24.84 & - & 29.75 & - \\
        QCN~\cite{Yang2019EnhancingCM} & - & - & 25.19 & - & 22.37 & - & 29.82 & - \\
        MCN~\cite{zheng2020toward} & 19.1 & 25.7 & 25.10 & 29.09 & 24.91 & 26.97 & 30.40 & 32.63 \\ 
        G-Trans~\cite{bao2021g} & - & - & 25.12 & 27.17 & \textbf{25.52} & \textbf{27.11} & \textbf{32.39} & 34.08 \\ \hline
        SR Doc2Doc & - & 8.62 & - & 4.70 & - & 21.18 & - & 34.16 \\
        MR Doc2Sent & \textbf{19.4} & 25.8 & \textbf{25.24} & 29.20 & 25.00 & 26.70 & 32.11 & 34.18 \\
        MR Doc2Doc & - & \textbf{25.9} & - & \textbf{29.27} & - & 26.71 & - & \textbf{34.48} \\ \hline \hline
        Sent2Sent ++ & 21.9 & 27.9 & 27.12 & 30.74 & 27.85 & 29.41 & 32.14 & 34.20 \\
        SR Doc2Doc ++ & - & 27.0 & - & 29.96 & - & 30.61 & - & 34.38 \\
        MR Doc2Sent ++ & \textbf{22.0} & 28.1 & \textbf{27.34} & 30.98 & \textbf{29.50} & 31.17 & \textbf{32.44} & 34.52 \\
        MR Doc2Doc ++ & - & \textbf{28.4} & - & \textbf{31.37} & - & \textbf{32.59} & - & \textbf{34.91} \\ \hline 
        
    \end{tabular}}
    \caption{Experiment results of document translation. ``-" means not provided. Except baseline cited from previous papers, we also re-implement our strong baseline with the best hyper-parameters (dropout, as is in section~\ref{sec:reg}) on the development sets.
    ``++'' indicates using additional sentence corpus. From the upper part, though SR Doc2Doc yields disappointing translation and even fails on \textit{TED}, MR Doc2Doc achieves much better results, proving the feasibility of Doc2Doc. From the lower part, extra sentence-level corpus can activate SR Doc2Doc and boost MR Doc2Doc, yielding the best results.}
    \label{tab:slct_mix}
\end{table*}

\section{Experiment Settings}

\subsection{Datasets}

For our main experiments, we follow the datasets provided by \citet{Maruf2019SelectiveAF} and \citet{zheng2020toward}, including \textit{TED} (ZH-EN/EN-DE), \textit{News} (EN-DE), and \textit{Europarl} (EN-DE). The Chinese-English and English-German TED datasets are from IWSLT 2015 and 2017 evaluation campaigns, respectively. For ZH-EN, we use dev2010 as the development set and tst2010-2013 as the test set. For TED (EN-DE), we use tst2016-2017 as the test set and the rest as the development set. For News (EN-DE). the training/develop/test sets are: News Commentary v11, WMT newstest2015, and WMT newstest2016. For Europarl (EN-DE). The corpus is extracted from the Europarl v7 according to the method proposed in~\citet{Maruf2019SelectiveAF}.~\footnote{EN-DE datasets are from~\url{https://github.com/sameenmaruf/selective-attn}}

Experiments on Spanish, French, Russian to English are also conducted, whose training sets are News Commentary v14
, with the development sets and test sets are newstest2012 / newstest2013 (ES-EN), newstest2013 / newstest2014 (FR-EN), newstest2018 / newstest2019 (RU-EN), respectively. 

Besides, two additional sentence-level datasets are also adopted. For EN-DE, we use \textit{Wikipedia}
, a corpus containing 2.4 million pairs of sentences. For ZH-EN, we extract one-tenth of WMT 2019
, around 2 million sentence pairs.
 
Additionally, a document-level dataset with contrastive test sets in EN-RU~\cite{Voita2019WhenAG} is used to evaluate lexical coherence.

Lastly, we propose a new document-level dataset in this paper, whose source, scales, and benchmark will be illustrated in the subsequent sections.

For sentences without any ending symbol inside documents, periods are manually added. 
For our Doc2Doc experiments, the development and test sets are documents merged by sentences. 
We list all the detailed information of used datasets in Table~\ref{tab:datasets}, including languages, scales, and downloading URLs for reproducibility.

\subsection{Models}

For the model setting, we follow the base version of Transformers~\cite{Vaswani2017AttentionIA}, including 6 layers for both encoders and decoders, 512 dimensions for model, 2048 dimensions for ffn layers, 8 heads for attention. For all experiments, we use subword~\cite{sennrich2016bpe} with 32K merge operations on both sides and cut out tokens appearing less than five times. The models are trained with a batch size of 32000 tokens on 8 Tesla V100 GPUs. Parameters are optimized by using Adam optimizer \cite{Kingma2015AdamAM}, with $\beta_1 = 0.9$, $\beta_2 = 0.98$, and $\epsilon = 10^{-9}$. The learning rate is scheduled according to the method proposed in \citet{Vaswani2017AttentionIA}, with $warmup\_steps = 4000$. Label smoothing \cite{Szegedy2016RethinkingTI} of value=0.1 is also adopted. We set dropout=0.3 for small datasets like \textit{TED} and \textit{News}, and dropout=0.1 for larger datasets like \textit{Europarl}, unless stated elsewise. 



\subsection{Evaluation}

For inference, we generate the hypothesis with a beam size of 5. Following previous related work, we adopt tokenized case-insensitive BLEU~\cite{Papineni2001BleuAM}. Specifically, we follow the methods in ~\citet{Liu2020MultilingualDP}, which calculate sentence-level BLEU (denoted as s-BLEU) and document-level BLEU (denoted as d-BLEU), respectively. For d-BLEU, the computing object is either the concatenation of generated sentences or the directly generated documents. Since our documents are generated atomically and hard to split into sentences, we only report d-BLEU for Doc2Doc.






\section{Results and Analysis}

\subsection{MR Doc2Doc Improves Performance}

\paragraph{MR matters.} It can be seen from the upper part of Table~\ref{tab:slct_mix} that SR Doc2Doc indeed has a severe drop on \textit{News} and even fails to generate normal results on \textit{TED}, which accords with the findings of ~\citet{Zhang2018ImprovingTT,Liu2020MultilingualDP}. It seems too hard to learn long-range translation directly.
However, once equipped with our training technique, MR Doc2Doc can yield the best results, outperforming our strong baseline and previous works on \textit{TED} and \textit{Europarl}. We suggest that NMT is able to acquire the capacity of translating long-range context, as long as it cooperates with some shorter segments as assistance. With the multi-resolutional help of easier patterns, the model can gradually master how to generate complicated sequences.

\paragraph{Doc2Doc matters.} We also compare MR Doc2Doc to a intuitive baseline: MR Doc2Sent. The latter one is trained in a typical Doc2Sent way: the source is the whole past context, the target is the current sentence. From the experimental results, we can see Doc2Doc outperforms it due to much broader contexts. Language model can effectively improve translation performance~\cite{sun2021multilingual}.


To show the universality of MR Doc2Doc, we also conduct the experiments on other language pairs: Spanish, French, Russian to English. As is shown in Table~\ref{tab:more_langs}, MR Doc2Doc can be achieved on all language pairs and obtains comparable or better results compared with Sent2Sent.

\begin{table}[ht]\footnotesize
    \centering
    \begin{tabular}{lccc}
        \hline
        Models & ES-EN & FR-EN & RU-EN \\ \hline
        Sent2Sent & \textbf{29.55} & 28.69 & 23.22 \\
        SR Doc2Doc & 26.79 & 23.86 & 16.47 \\
        MR Doc2Sent & 29.23 & 28.75 & 23.48 \\
        MR Doc2Doc & 29.37 & \textbf{28.85} & \textbf{23.98} \\ \hline
    \end{tabular}
    \caption{Document translation experiments on more languages, showing the comprehensive effectiveness.}
    \label{tab:more_langs}
\end{table}

It is worth noting that all our results are obtained without any adjustment of model architecture or any extra parameters.

\subsection{Additional Sentence Corpus Helps}
\label{add_sentences}

Furthermore, introducing extra sentence-level corpus is also an effective technique. This can be regarded as another form of multi-resolutional training, as it supplements more sentence-level information. This strategy makes an impact in two ways: activating SR Doc2Doc and boosting MR Doc2Doc.

We merge the datasets mentioned above and \textit{Wikipedia} (EN-DE), WMT (ZH-EN), two out-of-domain sentence-level datasets to do experiments. \footnote{Sentences and documents in non-MR settings are oversampled for six times to keep the same data ratio with the MR settings, 
which is proved helpful to the performance in Appendix~\ref{appendix:os}.
Due to the larger scale, we find the settings of dropout=0.2 for \textit{TED}, \textit{News} and dropout=0.1 for \textit{Europarl} yield the best results for both Sent2Sent and Doc2Doc.}


As is shown in the lower part of Table~\ref{tab:slct_mix}, on the one hand, SR Doc2Doc models are activated and can reach comparable levels with Sent2Sent models as long as assisted with additional sentences. On the other hand, MR Doc2Doc obtains the best results on all datasets and further widens the gap with the sentence corpus's boost. Even out-of-domain sentences can leverage the learning ability of document translation. It again proves the importance of multi-resolutional assistance.

In addition, as analyzed in the previous section, Doc2Sent models are not compatible with sentence-level corpus since the model entrance is specially designed for consecutive sentences. However, Doc2Doc models can naturally draw on the merits of any parallel pairs, including piecemeal sentences. Considering the amount of parallel sentence-level data is much larger than the document-level one, MR Doc2Doc has a powerful application potential compared with Doc2Sent.



\subsection{Further Analysis on MR Doc2Doc}

\subsubsection{Improved Discourse Coherence}

Except for BLEU, whether Doc2Doc truly learns to utilize the context to resolve discourse inconsistencies has to be verified. We use the contrastive test sets proposed by~\citet{Voita2019WhenAG}, which include deixis, lexicon consistency, ellipsis (inflection), and ellipsis (verb phrase) on English-Russian. Each instance contains a positive translation and a few negative ones, whose difference is only one specific word. With force decoding, if the score of the positive one is the highest, then this instance is counted as correct. 

As is shown in Table~\ref{tab:voita}, MR Doc2Doc achieves significant improvements and obtain the best results, which proves MR Doc2Doc indeed well captures the context information and maintain the cross-sentence coherence.

\begin{table}[ht]\footnotesize
    \centering
    \begin{tabular}{lcccc}
        \hline
        Models & deixis & lex.c & ell.infl & ell.VP \\ \hline
        Sent2Sent & 51.1 & 45.6 & 55.4 & 27.4 \\
        \citet{zheng2020toward} & 61.3 & 46.1 & 61.0 & 35.6 \\
        MR Doc2Doc & \textbf{64.7} & \textbf{46.3} & \textbf{65.9} & \textbf{53.0} \\ \hline
    \end{tabular}
    \caption{Discourse phenomena evaluation on the contrastive test sets. Our Doc2Doc shows a much better capacity for building the document coherence.}
    \label{tab:voita}
\end{table}

\subsubsection{Strong Context Sensibility}

\citet{li2020does} find the performance of previous context-aware systems does not decrease with intentional incorrect context and suspect the context usage of context encoders. To verify whether Doc2Doc truly takes advantage of the contextual information in the document, we also conduct the inference with the wrong context deliberately. If the model neglects discourse dependency, then there should be no difference in the performance.

Specifically, we firstly shuffle the sentence order inside each document randomly, marking it as \textit{Local Shuffle}. Furthermore, we randomly swap sentences among all the documents to make the context more disordered, marking it as \textit{Global Shuffle}. As is shown in Table~\ref{tab:shuffle}, the misleading context results in a significant drop for the Doc2Doc model in BLEU. Besides, Global Shuffle brings more harm than Local Shuffle, showing that more chaotic contexts lead to more harm. After all, Local Shuffle still reserves some general information, like topic or tense. These experiments 
prove the usage of the context.

\begin{table}[ht]\footnotesize
    \centering
    \begin{tabular}{l|c|ccc}
        \hline
        \multirow{2}{*}{Models} & ZH-EN & \multicolumn{3}{c}{EN-DE} \\
         & TED & TED & News & Europarl \\ \hline
        MR Doc2Doc & 25.84 & 29.27 & 26.71 & 34.48 \\ \hline
        Local Shuffle & 24.10 & 27.48 & 25.22 & 33.52 \\
        Global Shuffle & 23.69 & 27.17 & 24.96 & 32.47 \\ \hline
    \end{tabular}
    \caption{Misleading contexts can bring negative effects to Doc2Doc, proving the dependent usage of the context information. And more chaotic contexts harm more (Global vs. Local).}
    \label{tab:shuffle}
\end{table}

\subsubsection{Compatible with Sentences}

\begin{figure}[htb]
    \centering
    \includegraphics[scale=0.45]{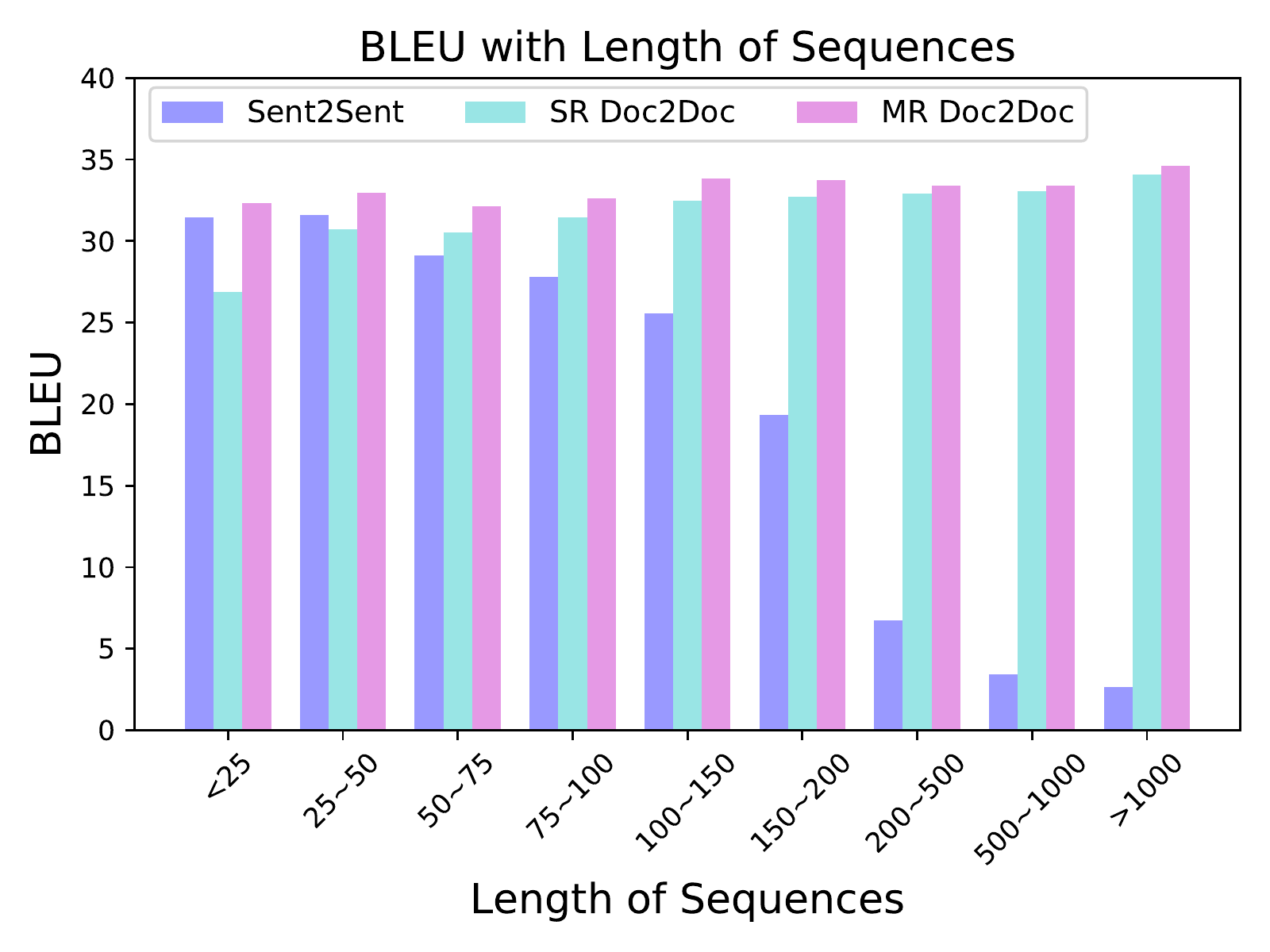}
    \captionsetup{font={footnotesize}}
    \caption{The model trained only on sentence-level or document-level corpus fails to translate sequences in unseen lengths while the MR model yields the best results in all scenarios.}
    \label{fig:bleu_with_length}
\end{figure}

The performance with sequence length is also analyzed in this study. Taking \textit{Europarl} as an example, we randomly split documents into shorter paragraphs in different lengths and evaluate them with our models, as is shown in Figure~\ref{fig:bleu_with_length}. Obviously, the model trained only on sentence-level corpus has a severe drop when translating long sequences, while the model trained only on document-level corpus shows the opposite result, which reveals the importance of data distribution. However, the model trained with our multi-resolutional strategy can sufficiently cope with all situations, breaking the limitation of sequence length in translation. By conducting MR Doc2Doc, we obtain an all-in-one model that is capable of translating sequences of any length, avoiding deploying two systems for sentences and documents, respectively.

\section{Further Evidence with Newly Proposed Datasets and Metrics}
\label{sec:datasets}




To further verify our conclusions and push the development of this field, we also contribute a new dataset along with new metrics.
Specifically, we propose a package of a large and diverse parallel document corpus, three deliberately designed metrics, and correspondingly constructed test sets~\footnote{\url{https://github.com/sunzewei2715/Doc2Doc_NMT}}. On the one hand, they make our conclusions more solid. On the other hand, they may benefit future researches to expand the comparison scenes.

\subsection{Parallel Document Corpus}

We crawl bilingual news corpus from two websites\footnote{\url{https://cn.nytimes.com}} \footnote{\url{https://cn.ft.com}} with both English and Chinese content provided. The detailed cleaning procedure is in Appendix~\ref{appendix:pdc}. Finally, $1.39$ million parallel sentences within almost $60$ thousand parallel documents are collected. The corpus contains large-scale data with internal dependency in different lengths and diverse domains, including politics, finance, health, culture, etc. 
We name it \textbf{PDC} (Parallel Document Corpus).

\subsection{Metrics}

To inspect the coherence improvement, we sum up three common linguistic features in document corpus that the Sent2Sent model can not handle:

\textbf{Tense Consistency (TC):} 
If the source language is tenseless (e.g. Chinese), it is hard for Sent2Sent models to maintain the consistency of tense.

\textbf{Conjunction Presence (CP):} 
Traditional models ignore cross-sentence dependencies, and the sentence-level translation may cause the missing of conjunctions like ``And'' \cite{Xiong2018ModelingCF}.

\textbf{Pronoun Translation (PT):} 
In pro-drop languages such as Chinese and Japanese, pronouns are frequently omitted. When translating from a pro-drop language into a non-pro-drop language (e.g., Chinese-to-English), invisible dropped pronouns may be missing \cite{Wang2016DroppedPG,Wang2016ANA,Wang2018TranslatingPL,Wang2018LearningTJ}. 

Afterward, we collect documents that contain abundant verbs in the past tense, conjunctions, and pronouns, as test sets. These words, as well as their positions, are labeled. 
Some cases are in Appendix~\ref{appendix:cases}. 

For each word-position pair $<w, p>$, we check whether $w$ appears in the generated documents within a rough span. And we calculate the appearance percentage as the evaluation score, Specifically:

\begin{equation}
      \text{TC / CP / PT} = \frac{\sum_{i}^{n}\sum_{j}^{|W_i|} \mathbb{I}(w_{ij} \in y_{i}^{\text{span}})}{\sum_{i}^{n}|W_i|}
      \label{eq:metric}
\end{equation}

\begin{equation}
      \text{span} = \left[ \alpha_i p_{ij} -d, \alpha_i p_{ij} +d \right]
\end{equation}


$n$ indicates the number of sequences in the test set, $W_i$ indicates the labeled word set of sequence$_i$, $w$ indicates labeled words, $y_i$ indicates output$_i$, $p_{ij}$ indicates the labeled position of $w_{ij}$ in the reference$_i$, $\alpha_i$ indicates the length ratio of translation and reference, $d$ indicates the span radius. We set $d=20$ in this paper, and calculate the geometric mean as the overall score denoted as \textbf{TCP}.

\subsection{Test Sets}
Along with the filtration of the aforementioned coherence indices, the test sets are built based on websites that are totally different from the training corpus to avoid overfitting. 
Meanwhile, to alleviate the bias of human translation, the English documents are selected as the reference and manually translated to the Chinese documents as the source. 
Finally, a total of nearly five thousand sentences within 148 documents is obtained.


\subsubsection{Benchmark}

Basic experiments with Sent2Sent and Doc2Doc are conducted based on our new datasets, along with full WMT ZH-EN corpus, a sentence-level dataset containing around 20 million pairs. 
\footnote{We set dropout=0.2 for Sent2Sent and MR Doc2Doc without WMT, and dropout=0.1 for the rest settings according to the performance on the development set. Oversampling is done again, as aforementioned, to enhance the performance for non-MR settings.} 
We use WMT newstest2019 as the development set and evaluate the models with our new test sets as well as metrics. The results are shown in Table~\ref{tab:bench}.

\begin{table}[ht]\scriptsize
    \centering
    \begin{tabular}{l|c|cccc|c}
        \hline
        Systems & d-BLEU & TC & CP & PT & TCP & Man\\ \hline
        Sent2Sent & 27.05 & 54.0 & 25.5 & 62.5 & 44.1 & 2.89\\
        SR Doc2Doc & 24.33 & 46.7 & 24.8 & 61.5 & 41.5 & 2.87 \\
        MR Doc2Doc & \textbf{27.80} & \textbf{56.9} & \textbf{25.7} & \textbf{63.9} & \textbf{45.4} & \textbf{3.02} \\
        \hline
        Sent2Sent ++ & 30.28 & 58.3 & 34.1 & 64.5 & 50.4 & 3.58 \\
        SR Doc2Doc ++ & 31.20 & 59.3 & 36.3 & 64.9 & 51.9 & 3.61 \\
        MR Doc2Doc ++ & \textbf{31.62} & \textbf{59.7} & \textbf{36.3} & \textbf{65.9} & \textbf{52.3} & \textbf{3.69} \\
        \hline
    \end{tabular}
    \caption{Benckmark of our new datasets. ``++'' indicates using additional WMT corpus. ``Man'' refers to human evaluation. Doc2Doc shows much better results in all terms.}
    \label{tab:bench}
\end{table}

\textbf{BLEU:} In terms of BLEU, MR Doc2Doc outperforms Sent2Sent, illustrating the positive effect of long-range context. Moreover, with extra sentence-level corpus, Doc2Doc shows significant improvements again. 

\textbf{Fine-grained Metrics:} Our metrics show much clearer improvements. Considering the usage of contextual information, tense consistency is better guaranteed with Doc2Doc. Meanwhile, Doc2Doc is much more capable of translating the invisible pronouns by capturing original referent beyond the current sentence. Finally, the conjunction presence shows the same tendency.

\textbf{Human Evaluation:} Human evaluation is also conducted to illustrate the reliability of our metrics. One-fifth of translated documents are sampled and scored by linguistics experts from 1 to 5 according to not only translation quality but also translation consistency~\cite{sun2020generating}. 
As is shown in Table~\ref{tab:bench}, human evaluation shows a strong correlation with TCP. More specifically, the Pearson Correlation Coefficient (PCCs) between human scores and TCP is higher than that of BLEU (97.9 vs. 94.1).



\subsection{Case Study}

Table~\ref{tab:case-coherence} shows an example of document translation. Sent2Sent model neglects the cross-sentence context and mistakenly translate the ambiguous word, which leads to a confusing reading experience. However, the Doc2Doc model can grasp a full picture of the historical context and make accurate decisions.

\begin{table}[!ht]\scriptsize
    \centering
    \begin{tabular}{p{0.15\columnwidth}|p{0.77\columnwidth}}
        \hline
        & \begin{CJK*}{UTF8}{gbsn}与大多数欧洲人一样, {\color{blue}德国}{\color{red}总理}对美国总统的“美国优先”民族主义难以掩饰不屑。
        \end{CJK*}\\
        Source& ... \\
        & \begin{CJK*}{UTF8}{gbsn} 但她已进入第四个、也必定是最后一个{\color{red}总理}任期 。\end{CJK*} \\
        \hline
        & Like most Europeans , the {\color{blue}German} {\color{red}chancellor} has struggled to hide his disdain for the US president’s “America First” nationalism.\\
        Sent2Sent & ... \\
         & But she has entered a fourth and surely last term as {\color{red}prime minister}. \\ \hline
         & Like most Europeans, the {\color{blue}German} {\color{red}chancellor}’s disdain for the US president’s “America First” nationalism is hard to hide. \\
        Doc2Doc & ... \\
         & But she has entered her fourth and certainly final term as {\color{red}chancellor}. \\ \hline
    \end{tabular}
    \captionsetup{font={footnotesize}}
    \caption{Coherence problem in document translation. Without discourse contexts, the Chinese word \begin{CJK*}{UTF8}{gbsn}``总理'' \end{CJK*} is usually translated to ``prime minister'', while in the context of ``German'', it should be translated into ``chancellor''. }
    \label{tab:case-coherence}
\end{table}

Also, we manually switch the context information in the source side to test the model sensibility, as is shown in Table~\ref{tab:case-country}. It turns out that Doc2Doc is able to adapt to different contexts.

\begin{table}[!ht]\scriptsize
    \centering
    \begin{tabular}{l|ccc}
        \hline
        Country & Sent2Sent & Doc2Doc & Oracle \\ \hline
        Germany & prime minister & \textbf{chancellor} & chancellor \\
        Italy & \textbf{prime minister} & \textbf{prime minister} & prime minister \\
        Austria & prime minister & \textbf{chancellor} & chancellor \\
        France & \textbf{prime minister} & \textbf{prime minister} & prime minister \\ \hline
        
    \end{tabular}
    \captionsetup{font={footnotesize}}
    \caption{Further study of Table~\ref{tab:case-coherence}. We switch the country information in the source side like \textit{German} $\rightarrow$ \textit{Italian}/\textit{Austrian}/\textit{French}, \textit{Berlin} $\rightarrow$ \textit{Rome}/\textit{Vienna}/\textit{Paris}. Doc2Doc model shows strong sensibility to the discourse context.}
    \label{tab:case-country}
\end{table}

\section{Limitation}

Though multi-resolutional Doc2Doc achieves direct document translation and obtains better results, there still exists a big challenge: efficiency. The computation cost of self-attention in Transformer rises with the square of the sequence length. As we feed the entire document into the model, the memory usage will be a bottleneck for larger model deployment. 
And the inference speed may be affected if no parallel operation is conducted.
Recently, many studies focus on the efficiency enhancement on long-range sequence processing~\cite{correia2019adaptively,child2019generating,kitaev2020reformer,wu2020lite,beltagy2020longformer,rae2020compressive}. We leave reducing the computation cost to the future work.

\section{Related Work}
\label{sec:related}

Document-level neural machine translation is an important task and has been abundantly studied with multiple datasets as well as methods.

The mainstream research in this field is the model architecture improvement. Specifically, several recent attempts extend the Sent2Sent approach to the Doc2Sent-like one. \citet{Wang2017ExploitingCC,Werlen2018DocumentLevelNM,Tan2019HierarchicalMO} make use of hierarchical RNNs or Transformer to summarize previous sentences.
\citet{Jean2017NeuralMT,Bawden2017EvaluatingDP,Zhang2018ImprovingTT,Voita2018ContextAwareNM,Kuang2018FusingRI,Maruf2019SelectiveAF,Yang2019EnhancingCM,Jiang2019DocumentlevelNM,zheng2020toward,yun2020improving,xuefficient} introduce additional encoders or query layers with attention model and feed the history contexts into decoders.
\citet{Maruf2018DocumentCN,Kuang2018ModelingCF,Tu2018LearningTR} propose to augment NMT models with a cache-like memory network, which generates the translation depending on the decoder history retrieved from the memory. 

Besides, some works intend to resolve this problem in other ways. 
\citet{Jean2019ContextAwareLF} propose a regularization term for encouraging to focus more on the additional context using a multi-level pair-wise ranking loss. 
\citet{Yu2020BetterDM} utilize a noisy channel reranker with Bayes' rule. 
\citet{Garcia2019ContextAwareNM} extends the beam search decoding process with fusing an attentional RNN with an SSLM by modifying the computation of the final score. 
\citet{Saunders2020UsingCI} present an approach for structured loss training with document-level objective functions. 
\citet{Liu2020MultilingualDP,ma2020simple} combine large-scale pre-train model with DNMT. 
\citet{Unanue2020LeveragingDR,Kang2020DynamicCS} adopt reinforcement learning methods. 

There are also some works sharing similar ideas with us. \citet{Tiedemann2017NeuralMT,Bawden2017EvaluatingDP} explore concatenating two consecutive sentences and generate two sentences directly. Obviously, we leverage greatly longer information and capture the full context.
\citet{JunczysDowmunt2019MicrosoftTA} cut documents into long segments and feed them into training like BERT~\cite{devlin2019bert}. There are at least three main differences. Firstly, they need to add specific boundary tokens between sentences while we directly translate the original documents without any additional processing. Secondly, we propose a novel multi-resolutional training paradigm that shows consistent improvements compared with regular training. Thirdly, for extremely long documents, they restrict the segment length to 1000 tokens or make a truncation while we preserve entire documents and achieve literal document-to-document training and inference.

Finally, our work is also related to a series of studies in long sequence generation like GPT \cite{Radford2018ImprovingLU}, GPT-2 \cite{Radford2019LanguageMA}, and Transformer-XL \cite{Dai2019TransformerXLAL}. We all suggest that the deep neural generation models have the potential to well process long-range sequences.

\section{Conclusion}
\label{sec:conclusion}

In this paper, we try to answer the question of whether Document-to-document translation works. It seems naive Doc2Doc can fail in multiple scenes. However, with the multi-resolutional training proposed in this paper, it can be successfully activated. Different from traditional methods of modifying the model architectures, our approach introduces no extra parameters. A comprehensive set of experiments on various metrics show the advantage of MR Doc2Doc. In addition, we contribute a new document-level dataset as well as three new metrics to the community.

\bibliography{acl}

\begin{thebibliography}{60}
\expandafter\ifx\csname natexlab\endcsname\relax\def\natexlab#1{#1}\fi

\bibitem[{Bahdanau et~al.(2015)Bahdanau, Cho, and
  Bengio}]{Bahdanau2015NeuralMT}
Dzmitry Bahdanau, Kyunghyun Cho, and Yoshua Bengio. 2015.
\newblock Neural machine translation by jointly learning to align and
  translate.
\newblock In \emph{ICLR}.

\bibitem[{Bao et~al.(2021)Bao, Zhang, Teng, Chen, and Luo}]{bao2021g}
Guangsheng Bao, Yue Zhang, Zhiyang Teng, Boxing Chen, and Weihua Luo. 2021.
\newblock G-transformer for document-level machine translation.
\newblock In \emph{ACL}.

\bibitem[{Bawden et~al.(2017)Bawden, Sennrich, Birch, and
  Haddow}]{Bawden2017EvaluatingDP}
Rachel Bawden, Rico Sennrich, Alexandra Birch, and Barry Haddow. 2017.
\newblock Evaluating discourse phenomena in neural machine translation.
\newblock In \emph{NAACL-HLT}.

\bibitem[{Beltagy et~al.(2019)Beltagy, Peters, and
  Cohan}]{beltagy2020longformer}
Iz~Beltagy, Matthew~E Peters, and Arman Cohan. 2019.
\newblock Longformer: The long-document transformer.
\newblock \emph{arXiv}, abs/2004.05150.

\bibitem[{Child et~al.(2019)Child, Gray, Radford, and
  Sutskever}]{child2019generating}
Rewon Child, Scott Gray, Alec Radford, and Ilya Sutskever. 2019.
\newblock Generating long sequences with sparse transformers.
\newblock \emph{arXiv}, abs/1904.10509.

\bibitem[{Correia et~al.(2019)Correia, Niculae, and
  Martins}]{correia2019adaptively}
Gon{\c{c}}alo~M Correia, Vlad Niculae, and Andr{\'e}~FT Martins. 2019.
\newblock Adaptively sparse transformers.
\newblock In \emph{EMNLP-IJCNLP}.

\bibitem[{Dai et~al.(2019)Dai, Yang, Yang, Carbonell, Le, and
  Salakhutdinov}]{Dai2019TransformerXLAL}
Zihang Dai, Zhilin Yang, Yiming Yang, Jaime~G. Carbonell, Quoc~V. Le, and
  Ruslan Salakhutdinov. 2019.
\newblock Transformer-xl: Attentive language models beyond a fixed-length
  context.
\newblock In \emph{ACL}.

\bibitem[{Devlin et~al.(2019)Devlin, Chang, Lee, and
  Toutanova}]{devlin2019bert}
Jacob Devlin, Ming-Wei Chang, Kenton Lee, and Kristina Toutanova. 2019.
\newblock Bert: Pre-training of deep bidirectional transformers for language
  understanding.
\newblock In \emph{NAACL-HLT}.

\bibitem[{Garcia et~al.(2019)Garcia, Creus, and
  Espa{\~n}a-Bonet}]{Garcia2019ContextAwareNM}
Eva~Mart{\'i}nez Garcia, C.~Creus, and C.~Espa{\~n}a-Bonet. 2019.
\newblock Context-aware neural machine translation decoding.
\newblock In \emph{DiscoMT@EMNLP}.

\bibitem[{Jean and Cho(2019)}]{Jean2019ContextAwareLF}
S{\'e}bastien Jean and Kyunghyun Cho. 2019.
\newblock Context-aware learning for neural machine translation.
\newblock \emph{arXiv}, abs/1903.04715.

\bibitem[{Jean et~al.(2017)Jean, Lauly, Firat, and Cho}]{Jean2017NeuralMT}
S{\'e}bastien Jean, Stanislas Lauly, Orhan Firat, and Kyunghyun Cho. 2017.
\newblock Neural machine translation for cross-lingual pronoun prediction.
\newblock In \emph{DiscoMT@EMNLP}.

\bibitem[{Jiang et~al.(2019)Jiang, Wang, Li, Utiyama, Chen, Sumita, Zhao, and
  Lu}]{Jiang2019DocumentlevelNM}
Shu Jiang, Rui Wang, Zuchao Li, Masao Utiyama, Kehai Chen, Eiichiro Sumita, Hai
  Zhao, and Bao-Liang Lu. 2019.
\newblock Document-level neural machine translation with inter-sentence
  attention.
\newblock \emph{arXiv}, abs/1910.14528.

\bibitem[{Junczys-Dowmunt(2019)}]{JunczysDowmunt2019MicrosoftTA}
Marcin Junczys-Dowmunt. 2019.
\newblock Microsoft translator at wmt 2019: Towards large-scale document-level
  neural machine translation.
\newblock In \emph{WMT}.

\bibitem[{Jwalapuram et~al.(2020)Jwalapuram, Rychalska, Joty, and
  Basaj}]{Jwalapuram2020CanYC}
Prathyusha Jwalapuram, Barbara Rychalska, Shafiq Joty, and Dominika Basaj.
  2020.
\newblock Can your context-aware mt system pass the dip benchmark tests? :
  Evaluation benchmarks for discourse phenomena in machine translation.
\newblock \emph{arXiv}, abs/2004.14607.

\bibitem[{Kang et~al.(2020)Kang, Zhao, Zhang, and Zong}]{Kang2020DynamicCS}
Xiaomian Kang, Yang Zhao, Jiajun Zhang, and Chengqing Zong. 2020.
\newblock Dynamic context selection for document-level neural machine
  translation via reinforcement learning.
\newblock In \emph{EMNLP}.

\bibitem[{Kim et~al.(2019)Kim, Tran, and Ney}]{Kim2019WhenAW}
Yunsu Kim, Thanh Tran, and Hermann Ney. 2019.
\newblock When and why is document-level context useful in neural machine
  translation?
\newblock In \emph{DiscoMT@EMNLP-IJCNLP}.

\bibitem[{Kingma and Ba(2015)}]{Kingma2015AdamAM}
Diederick~P Kingma and Jimmy Ba. 2015.
\newblock Adam: A method for stochastic optimization.
\newblock In \emph{ICLR}.

\bibitem[{Kitaev et~al.(2020)Kitaev, Kaiser, and Levskaya}]{kitaev2020reformer}
Nikita Kitaev, Lukasz Kaiser, and Anselm Levskaya. 2020.
\newblock Reformer: The efficient transformer.
\newblock In \emph{ICLR}.

\bibitem[{Koehn and Knowles(2017)}]{Koehn2017SixCF}
Philipp Koehn and Rebecca Knowles. 2017.
\newblock Six challenges for neural machine translation.
\newblock In \emph{NMT@ACL}.

\bibitem[{Kuang and Xiong(2018)}]{Kuang2018FusingRI}
Shaohui Kuang and Deyi Xiong. 2018.
\newblock Fusing recency into neural machine translation with an inter-sentence
  gate model.
\newblock In \emph{COLING}.

\bibitem[{Kuang et~al.(2018)Kuang, Xiong, Luo, and Zhou}]{Kuang2018ModelingCF}
Shaohui Kuang, Deyi Xiong, Weihua Luo, and Guodong Zhou. 2018.
\newblock Modeling coherence for neural machine translation with dynamic and
  topic caches.
\newblock In \emph{COLING}.

\bibitem[{L{\"a}ubli et~al.(2018)L{\"a}ubli, Sennrich, and
  Volk}]{laubli2018has}
Samuel L{\"a}ubli, Rico Sennrich, and Martin Volk. 2018.
\newblock Has machine translation achieved human parity? a case for
  document-level evaluation.
\newblock In \emph{EMNLP}.

\bibitem[{Li et~al.(2020)Li, Liu, Wang, Jiang, Xiao, Zhu, Liu, and
  Li}]{li2020does}
Bei Li, Hui Liu, Ziyang Wang, Yufan Jiang, Tong Xiao, Jingbo Zhu, Tongran Liu,
  and Changliang Li. 2020.
\newblock Does multi-encoder help? a case study on context-aware neural machine
  translation.
\newblock In \emph{ACL}.

\bibitem[{Liu et~al.(2020)Liu, Gu, Goyal, Li, Edunov, Ghazvininejad, Lewis, and
  Zettlemoyer}]{Liu2020MultilingualDP}
Yinhan Liu, Jiatao Gu, Naman Goyal, Xiongmin Li, Sergey Edunov, Marjan
  Ghazvininejad, Mike Lewis, and Luke Zettlemoyer. 2020.
\newblock Multilingual denoising pre-training for neural machine translation.
\newblock \emph{TACL}.

\bibitem[{Ma et~al.(2020)Ma, Zhang, and Zhou}]{ma2020simple}
Shuming Ma, Dongdong Zhang, and Ming Zhou. 2020.
\newblock A simple and effective unified encoder for document-level machine
  translation.
\newblock In \emph{ACL}.

\bibitem[{Maruf and Haffari(2018)}]{Maruf2018DocumentCN}
Sameen Maruf and Gholamreza Haffari. 2018.
\newblock Document context neural machine translation with memory networks.
\newblock In \emph{ACL}.

\bibitem[{Maruf et~al.(2019)Maruf, Martins, and Haffari}]{Maruf2019SelectiveAF}
Sameen Maruf, Andr{\'e} F.~T. Martins, and Gholamreza" Haffari. 2019.
\newblock Selective attention for context-aware neural machine translation.
\newblock In \emph{NAACL-HLT}.

\bibitem[{Miculicich et~al.(2018)Miculicich, Ram, Pappas, and
  Henderson}]{Werlen2018DocumentLevelNM}
Lesly Miculicich, Dhananjay Ram, Nikolaos Pappas, and James Henderson. 2018.
\newblock Document-level neural machine translation with hierarchical attention
  networks.
\newblock In \emph{EMNLP}.

\bibitem[{M{\"u}ller et~al.(2018)M{\"u}ller, Gonzales, Voita, and
  Sennrich}]{Mller2018ALT}
Mathias M{\"u}ller, Annette~Rios Gonzales, Elena Voita, and Rico Sennrich.
  2018.
\newblock A large-scale test set for the evaluation of context-aware pronoun
  translation in neural machine translation.
\newblock In \emph{WMT}.

\bibitem[{Papineni et~al.(2002)Papineni, Roukos, Ward, and
  Zhu}]{Papineni2001BleuAM}
Kishore Papineni, Salim Roukos, Todd Ward, and Wei-Jing Zhu. 2002.
\newblock Bleu: a method for automatic evaluation of machine translation.
\newblock In \emph{ACL}.

\bibitem[{Radford(2018)}]{Radford2018ImprovingLU}
Alec Radford. 2018.
\newblock Improving language understanding by generative pre-training.

\bibitem[{Radford et~al.(2019)Radford, Wu, Child, Luan, Amodei, and
  Sutskever}]{Radford2019LanguageMA}
Alec Radford, Jeffrey Wu, Rewon Child, David Luan, Dario Amodei, and Ilya
  Sutskever. 2019.
\newblock Language models are unsupervised multitask learners.

\bibitem[{Rae et~al.(2020)Rae, Potapenko, Jayakumar, Hillier, and
  Lillicrap}]{rae2020compressive}
Jack~W Rae, Anna Potapenko, Siddhant~M Jayakumar, Chloe Hillier, and Timothy~P
  Lillicrap. 2020.
\newblock Compressive transformers for long-range sequence modelling.
\newblock In \emph{ICLR}.

\bibitem[{Saunders et~al.(2020)Saunders, Stahlberg, and
  Byrne}]{Saunders2020UsingCI}
Danielle Saunders, Felix Stahlberg, and Bill Byrne. 2020.
\newblock Using context in neural machine translation training objectives.
\newblock In \emph{ACL}.

\bibitem[{Sennrich et~al.(2016)Sennrich, Haddow, and Birch}]{sennrich2016bpe}
Rico Sennrich, Barry Haddow, and Alexandra Birch. 2016.
\newblock Neural machine translation of rare words with subword units.
\newblock In \emph{ACL}.

\bibitem[{Sun et~al.(2020)Sun, Huang, Wei, Dai, and Chen}]{sun2020generating}
Zewei Sun, Shujian Huang, Hao-Ran Wei, Xin-yu Dai, and Jiajun Chen. 2020.
\newblock Generating diverse translation by manipulating multi-head attention.
\newblock In \emph{AAAI}.

\bibitem[{Sun et~al.(2021)Sun, Wang, and Li}]{sun2021multilingual}
Zewei Sun, Mingxuan Wang, and Lei Li. 2021.
\newblock Multilingual translation via grafting pre-trained language models.
\newblock In \emph{EMNLP-Findings}.

\bibitem[{Szegedy et~al.(2016)Szegedy, Vanhoucke, Ioffe, Shlens, and
  Wojna}]{Szegedy2016RethinkingTI}
Christian Szegedy, Vincent Vanhoucke, Sergey Ioffe, Jonathon Shlens, and
  Zbigniew Wojna. 2016.
\newblock Rethinking the inception architecture for computer vision.
\newblock In \emph{CVPR}.

\bibitem[{Tan et~al.(2019)Tan, Zhang, Xiong, and Zhou}]{Tan2019HierarchicalMO}
Xin Tan, Longyin Zhang, Deyi Xiong, and Guodong Zhou. 2019.
\newblock Hierarchical modeling of global context for document-level neural
  machine translation.
\newblock In \emph{EMNLP-IJCNLP}.

\bibitem[{Tiedemann and Scherrer(2017)}]{Tiedemann2017NeuralMT}
J{\"o}rg Tiedemann and Yves Scherrer. 2017.
\newblock Neural machine translation with extended context.
\newblock In \emph{DiscoMT@EMNLP}.

\bibitem[{Tu et~al.(2018)Tu, Liu, Shi, and Zhang}]{Tu2018LearningTR}
Zhaopeng Tu, Yang~P. Liu, Shuming Shi, and Tong Zhang. 2018.
\newblock Learning to remember translation history with a continuous cache.
\newblock \emph{TACL}.

\bibitem[{Unanue et~al.(2020)Unanue, Esmaili, Haffari, and
  Piccardi}]{Unanue2020LeveragingDR}
Inigo~Jauregi Unanue, Nazanin Esmaili, Gholamreza Haffari, and Massimo
  Piccardi. 2020.
\newblock Leveraging discourse rewards for document-level neural machine
  translation.
\newblock In \emph{COLING}.

\bibitem[{Vaswani et~al.(2017)Vaswani, Shazeer, Parmar, Uszkoreit, Jones,
  Gomez, Kaiser, and Polosukhin}]{Vaswani2017AttentionIA}
Ashish Vaswani, Noam Shazeer, Niki Parmar, Jakob Uszkoreit, Llion Jones,
  Aidan~N. Gomez, Lukasz Kaiser, and Illia Polosukhin. 2017.
\newblock Attention is all you need.
\newblock In \emph{NIPS}.

\bibitem[{Voita et~al.(2019)Voita, Sennrich, and Titov}]{Voita2019WhenAG}
Elena Voita, Rico Sennrich, and Ivan Titov. 2019.
\newblock When a good translation is wrong in context: Context-aware machine
  translation improves on deixis, ellipsis, and lexical cohesion.
\newblock In \emph{ACL}.

\bibitem[{Voita et~al.(2018)Voita, Serdyukov, Sennrich, and
  Titov}]{Voita2018ContextAwareNM}
Elena Voita, Pavel Serdyukov, Rico Sennrich, and Ivan Titov. 2018.
\newblock Context-aware neural machine translation learns anaphora resolution.
\newblock In \emph{ACL}.

\bibitem[{Wang et~al.(2018{\natexlab{a}})Wang, Tu, Shi, Zhang, Graham, and
  Liu}]{Wang2018TranslatingPL}
Longyue Wang, Zhaopeng Tu, Shuming Shi, Tong Zhang, Yvette Graham, and Qun Liu.
  2018{\natexlab{a}}.
\newblock Translating pro-drop languages with reconstruction models.
\newblock In \emph{AAAI}.

\bibitem[{Wang et~al.(2017)Wang, Tu, Way, and Liu}]{Wang2017ExploitingCC}
Longyue Wang, Zhaopeng Tu, Andy Way, and Qun Liu. 2017.
\newblock Exploiting cross-sentence context for neural machine translation.
\newblock In \emph{EMNLP}.

\bibitem[{Wang et~al.(2018{\natexlab{b}})Wang, Tu, Way, and
  Liu}]{Wang2018LearningTJ}
Longyue Wang, Zhaopeng Tu, Andy Way, and Qun Liu. 2018{\natexlab{b}}.
\newblock Learning to jointly translate and predict dropped pronouns with a
  shared reconstruction mechanism.
\newblock In \emph{EMNLP}.

\bibitem[{Wang et~al.(2016{\natexlab{a}})Wang, Tu, Zhang, Li, Way, and
  Liu}]{Wang2016ANA}
Longyue Wang, Zhaopeng Tu, Xiaojun Zhang, Hang Li, Andy Way, and Qun Liu.
  2016{\natexlab{a}}.
\newblock A novel approach to dropped pronoun translation.
\newblock In \emph{NAACL-HLT}.

\bibitem[{Wang et~al.(2016{\natexlab{b}})Wang, Zhang, Tu, Li, and
  Liu}]{Wang2016DroppedPG}
Longyue Wang, Xiaojun Zhang, Zhaopeng Tu, Hang Li, and Qun Liu.
  2016{\natexlab{b}}.
\newblock Dropped pronoun generation for dialogue machine translation.
\newblock In \emph{ICASSP}.

\bibitem[{Werlen and Popescu-Belis(2017)}]{Werlen2017ValidationOA}
Lesly~Miculicich Werlen and Andrei Popescu-Belis. 2017.
\newblock Validation of an automatic metric for the accuracy of pronoun
  translation (apt).
\newblock In \emph{DiscoMT@EMNLP}.

\bibitem[{Wu et~al.(2016)Wu, Schuster, Chen, Le, Norouzi, Macherey, Krikun,
  Cao, Gao, Macherey, Klingner, Shah, Johnson, Liu, Kaiser, Gouws, Kato, Kudo,
  Kazawa, Stevens, Kurian, Patil, Wang, Young, Smith, Riesa, Rudnick, Vinyals,
  Corrado, Hughes, and Dean}]{Wu2016GooglesNM}
Yonghui Wu, Mike Schuster, Zhifeng Chen, Quoc~V. Le, Mohammad Norouzi, Wolfgang
  Macherey, Maxim Krikun, Yuan Cao, Qin Gao, Klaus Macherey, Jeff Klingner,
  Apurva Shah, Melvin Johnson, Xiaobing Liu, Lukasz Kaiser, Stephan Gouws,
  Yoshikiyo Kato, Taku Kudo, Hideto Kazawa, Keith Stevens, George Kurian,
  Nishant Patil, Wei Wang, Cliff Young, Jason Smith, Jason Riesa, Alex Rudnick,
  Oriol Vinyals, Gregory~S. Corrado, Macduff Hughes, and Jeffrey Dean. 2016.
\newblock Google's neural machine translation system: Bridging the gap between
  human and machine translation.
\newblock \emph{arXiv}, abs/1609.08144.

\bibitem[{Wu et~al.(2020)Wu, Liu, Lin, Lin, and Han}]{wu2020lite}
Zhanghao Wu, Zhijian Liu, Ji~Lin, Yujun Lin, and Song Han. 2020.
\newblock Lite transformer with long-short range attention.
\newblock In \emph{ICLR}.

\bibitem[{Xiong et~al.(2018)Xiong, He, Wu, and Wang}]{Xiong2018ModelingCF}
Hao Xiong, Zhongjun He, Hua Wu, and Haifeng Wang. 2018.
\newblock Modeling coherence for discourse neural machine translation.
\newblock In \emph{AAAI}.

\bibitem[{Xu et~al.(2020)Xu, Xiong, van Genabith, and Liu}]{xuefficient}
Hongfei Xu, Deyi Xiong, Josef van Genabith, and Qiuhui Liu. 2020.
\newblock Efficient context-aware neural machine translation with layer-wise
  weighting and input-aware gating.
\newblock In \emph{DiscoMT@IJCAI}.

\bibitem[{Yang et~al.(2019)Yang, Zhang, Meng, Gu, Feng, and
  Zhou}]{Yang2019EnhancingCM}
Zhengxin Yang, Jinchao Zhang, Fandong Meng, Shuhao Gu, Yang Feng, and Jie Zhou.
  2019.
\newblock Enhancing context modeling with a query-guided capsule network for
  document-level nmt.
\newblock In \emph{EMNLP-IJCNLP}.

\bibitem[{Yu et~al.(2020)Yu, Sartran, Stokowiec, Ling, Kong, Blunsom, and
  Dyer}]{Yu2020BetterDM}
Lei Yu, Laurent Sartran, Wojciech Stokowiec, Wang Ling, Lingpeng Kong, Phil
  Blunsom, and Chris Dyer. 2020.
\newblock Better document-level machine translation with bayes' rule.
\newblock \emph{TACL}.

\bibitem[{Yun et~al.(2020)Yun, Hwang, and Jung}]{yun2020improving}
Hyeongu Yun, Yongkeun Hwang, and Kyomin Jung. 2020.
\newblock Improving context-aware neural machine translation using
  self-attentive sentence embedding.
\newblock In \emph{AAAI}.

\bibitem[{Zhang et~al.(2018)Zhang, Luan, Sun, Zhai, Xu, Zhang, and
  Liu}]{Zhang2018ImprovingTT}
Jiacheng Zhang, Huanbo Luan, Maosong Sun, Feifei Zhai, Jingfang Xu, Min Zhang,
  and Yang~P. Liu. 2018.
\newblock Improving the transformer translation model with document-level
  context.
\newblock In \emph{EMNLP}.

\bibitem[{Zheng et~al.(2020)Zheng, Yue, Huang, Chen, and
  Birch}]{zheng2020toward}
Zaixiang Zheng, Xiang Yue, Shujian Huang, Jiajun Chen, and Alexandra Birch.
  2020.
\newblock Toward making the most of context in neural machine translation.
\newblock In \emph{IJCAI}.

\end{thebibliography}
\bibliographystyle{acl_natbib}

\clearpage
\appendix

\section{Oversampling Illustration}
\label{appendix:os}

When combining document-level datasets with sentence-level datasets (especially out-of-domain corpus), we employ oversampling for non-MR settings. This can keep them the same data ratio with the MR setting and is helpful for their performance. Since the data size of MR is around 6 times of non-MR ($\approx \log_2 64$), as shown in Table~\ref{tab:ratio}, we mainly oversample for 6 times. The contrastive experiments are in Table~\ref{tab:os}. We attribute the improvements to the reduction of the proportion of out-of-domain data.

\begin{table}[ht]\footnotesize
    \centering
    \begin{tabular}{lc}
        \hline
        Datasets & Ratio \\ \hline
        TED (ZH-EN) & 6.7 \\
        TED (EN-DE) & 7.6 \\
        News (EN-DE) & 5.9 \\
        Europal & 4.6 \\
        News (ES-EN) & 5.9 \\
        News (FR-EN) & 5.9 \\
        News (RU-EN) & 5.9 \\
        PDC & 5.3 \\ \hline
        Mean & 6.0 \\ \hline
    \end{tabular}
    \caption{Ratio of MR/non-MR in data size}
    \label{tab:ratio}
\end{table}

\begin{table}[ht]\scriptsize
    \centering
    \begin{tabular}{lcccc}
        \hline
        \multirow{2}{*}{Dataset} & \multicolumn{2}{c}{Sent2Sent}  & \multicolumn{2}{c}{SR Doc2Doc}  \\
         & non-OS & OS & non-OS & OS \\ \hline
         TED(ZH-EN)+WMT & 27.52 & \textbf{27.90} & 26.05 & \textbf{26.67} \\
         TED(EN-DE)+Wiki & 29.19 & \textbf{30.74} & 29.81 & \textbf{29.96} \\
         News+Wiki & 27.77 & \textbf{29.41} & 30.15 & \textbf{30.61} \\
         Europarl+Wiki & 33.93 & \textbf{34.20} & 34.25 & \textbf{34.38}  \\
         PDC+WMT & 29.52 & \textbf{30.28} & 29.60 & \textbf{31.20}  \\
         \hline
    \end{tabular}
    \caption{The contrastive results of oversampling when combining sentence-level corpus.}
    \label{tab:os}
\end{table}



\section{Clean Procedure on PDC}
\label{appendix:pdc}

We mainly crawl bilingual news corpus from two websites~(\url{https://cn.nytimes.com}, \url{https://cn.ft.com}) with both English and Chinese content provided. Then three steps are followed to clean the corpus.

\begin{compactenum}
    \item \textbf{Deduplication:} We deduplicate the documents that include almost the same content.
    
    \item \textbf{Sentence Segmentation:} We use \textit{Pragmatic Segmenter} \footnote{\url{https://github.com/diasks2/pragmatic_segmenter}} to segment paragraphs into sentences.
    
    \item \textbf{Filtration:} We use \textit{fast\_align} \footnote{\url{https://github.com/clab/fast_align}} to align sentence pairs and label the pairs as misaligned ones if the alignment scores are less than $40\%$. Documents are finally removed if they contain misaligned sentence pairs.
\end{compactenum}
Finally, we obtain $1.39$ million parallel sentences within almost $60$ thousand cleaned parallel documents. The dataset contains diverse domains including politics, finance, health, culture, etc. 

\section{Cases of Our Test Sets}
\label{appendix:cases}

Apart from the statistic number in the main paper, we also provide some cases in our test sets to illustrate the value of our test sets and metrics, as shown in Table~\ref{tab:case-tense},\ref{tab:case-conjunction},\ref{tab:case-pronoun}.

\begin{table}[ht]\scriptsize
    \centering
    \begin{tabular}{l|p{0.8\columnwidth}}
        \hline
        Src & \begin{CJK*}{UTF8}{gbsn}1.双方 在 2017 年 都 向 法庭 提交 了 申请 。
        \end{CJK*}\\
           & \begin{CJK*}{UTF8}{gbsn} 2.邓普顿 奈特 {\color{red}想要} 报销 他 的 租金 。\end{CJK*} \\
           & \begin{CJK*}{UTF8}{gbsn} 3.伯德特 {\color{red}想要} 赶走 邓普顿 奈特 。\end{CJK*} \\
        \hline
        Ref & 1.Both parties had lodged applications with the tribunal in 2017. \\
          & 2.Templeton-Knight {\color{red}wanted} his rent reimbursed. \\
          & 3.Burdett {\color{red}wanted} to evict Templeton-Knight. \\
        \hline 
        NMT & 1.Both parties filed applications with the court in 2017. \\
          & 2.Templeton Knight {\color{red}wants} to reimburse his rent. \\
          & 3.Burdett {\color{red}wants} to get rid of Templeton Knight. \\ \hline
    \end{tabular}
    \captionsetup{font={footnotesize}}
    \caption{Tense inconsistency problem in translating tenseless languages (e.g. Chinese) to tense-sensitive languages (e.g. English). Individual sentences are translated into present tense with sentence-level models while the history context has provided the signal of past tense.}
    \label{tab:case-tense}
\end{table}

\begin{table}[ht]\scriptsize
    \centering
    \begin{tabular}{l|p{0.8\columnwidth}}
        \hline
        Src & \begin{CJK*}{UTF8}{gbsn}1.我 女儿 使用 的 胰岛素 类型 — — 世界 上 只有 两家 类似 类型 的 制造商 。\end{CJK*}\\
            & \begin{CJK*}{UTF8}{gbsn} 2.他们 继续 保持一致 同时 提高 价格 。\end{CJK*} \\
        \hline
        Ref & 1.The type of insulin that my daughter uses — there are only two manufacturers worldwide of a similar type. \\
          & 2.{\color{red}And} they continue to increase their prices lockstep together. \\
        \hline 
        NMT & 1.The type of insulin my daughter uses - there are only two manufacturers of similar types in the world. \\
          & 2.{\color{red}[conj miss]} They continue to be consistent while raising prices. \\ \hline
    \end{tabular}
    \captionsetup{font={footnotesize}}
    \caption{Conjunction missing problem in sentence-level translation. The sentences has strong semantic connection but are translated without any conjunction.}
    \label{tab:case-conjunction}
\end{table}

\begin{table}[ht]\scriptsize
    \centering
    \begin{tabular}{l|p{0.8\columnwidth}}
        \hline
        Src & \begin{CJK*}{UTF8}{gbsn}1.根据 市政府 的 说法 ， 奥特里 工厂 的 其他 拟议 功能 似乎 极 不 可能 实施 。\end{CJK*}\\
            & \begin{CJK*}{UTF8}{gbsn} 2.即使 顾问 和 调查人 推荐 {\color{red}[pro drop]}。\end{CJK*} \\
        \hline
        Ref & 1.Other proposed features for Autrey Mill seem highly unlikely to be implemented according to the City Manager. \\
          & 2.Even though consultants and surveys recommended {\color{red}them}. \\
        \hline 
        NMT$_A$ & 1.According to the city government, other proposed functions at the Autry plant appear highly unlikely to be implemented. \\
          & 2.Even if consultants and surveys recommend {\color{red}[pro miss]}. \\ \hline
        NMT$_B$ & 1.According to the municipal government , other proposed functions of the Autry plant seem highly impossible to implement . \\
          & 2.Even if consultants and surveys recommended {\color{red}it}. \\ \hline
    \end{tabular}
    \captionsetup{font={footnotesize}}
    \caption{Pronoun drop problem in translating pro-drop languages (e.g. Chinese) to non-pro-drop languages (e.g. English). The pronoun is omitted or translated wrongly with sentence-level models..}
    \label{tab:case-pronoun}
\end{table}

\end{document}